# High-dimensional Joint Sparsity Random Effects Model for Multi-task Learning

**Krishnakumar Balasubramanian**
Georgia Institute of Technology
krishnakumar3@gatech.edu

**Kai Yu**
Baidu Inc.
yukai@baidu.com

**Tong Zhang**
Rutgers University
tzhang@stat.rutgers.edu

## Abstract

Joint sparsity regularization in multi-task learning has attracted much attention in recent years. The traditional convex formulation employs the group Lasso relaxation to achieve joint sparsity across tasks. Although this approach leads to a simple convex formulation, it suffers from several issues due to the looseness of the relaxation. To remedy this problem, we view jointly sparse multi-task learning as a specialized random effects model, and derive a convex relaxation approach that involves two steps. The first step learns the covariance matrix of the coefficients using a convex formulation which we refer to as sparse covariance coding; the second step solves a ridge regression problem with a sparse quadratic regularizer based on the covariance matrix obtained in the first step. It is shown that this approach produces an asymptotically optimal quadratic regularizer in the multitask learning setting when the number of tasks approaches infinity. Experimental results demonstrate that the convex formulation obtained via the proposed model significantly outperforms group Lasso (and related multi-stage formulations).

## 1 Introduction

Modern high-dimensional data sets, typically with more parameters to estimate than the number of samples available, have triggered a flurry of research based on structured sparse models, both on the statistical and computational aspects. The initial problem considered in this setting was to estimate a sparse vector under a linear model (or the Lasso problem). Recently, several approaches have been proposed for estimating a sparse vector under additional constraints, for e.g., group sparsity- where certain groups of coefficients are jointly zero or non-zero. Another closely related problem is that of multi-task learning or simultaneous sparse approximation, which are special cases of the group sparse formulation. A de-facto procedure for dealing with joint sparsity regularization is the group-Lasso estimator [16], which is based on a $(2, 1)$-mixed norm convex relaxation to the non-convex $(2, 0)$-mixed norm formulation.

However, as we shall argue in this paper, group-Lasso suffers from several drawbacks due to the looseness of the relaxation; cf., [12, 9]. We propose a general method for multi-task learning in high-dimensions based on a joint sparsity random effects model. The standard approach for dealing with random effects requires estimating covariance information. Similarly, our estimation procedure involves two-steps: a convex covariance estimation step followed by the standard ridge-regression. The first step corresponds to estimating the covariance of the coefficients under additional constraints that promote sparsity. The intuition is that to deal with group sparsity (even if we are interested in estimating the coefficients) it is better to first estimate covariance information, and then plug in the covariance estimate for estimating the coefficients. With a particular sparse diagonal structure for the covariance matrix the model becomes similar to group-lasso, and the advantage of the proposed estimation approach over group-lasso formulation will be clarified in this setting.

**Related work:** Traditional estimation approaches for random effects model involve two-steps: first estimate the underlying covariance matrix, and then estimate the coefficients based on the covariance matrix. However, the traditional covariance estimation procedures are non-convex such as the popular method of restricted maximum likelihood (*REML*) and such models are typically studied in the low-dimensional setting [10].

From a Bayesian perspective, a hierarchical model for simultaneous sparse approximation is proposed in [15] based on a straightforward extension of automatic relevance determination. Under that setting, the tasks share a common hyper-prior that is estimated from the data by integrating out the actual parameter. The resulting marginal likelihood is maximized for the hyper-prior parameters; this proce-dure is called as type-II maximum likelihood in the literature. The non-Bayesian counterpart is called *random effects model* in classical statistics, and the resulting estimator is referred to as REML. The disadvantage of this approach is that it makes the resulting optimization problem non-convex and difficult to solve efficiently, as mentioned before. In addition, the problem becomes harder to analyze and provide convincing statistical and computational guarantees, while Lasso-related formulations are well studied and favorable statistical and computational properties could be established.

More recently, the problem of joint sparsity regularization has been studied under various settings (multi-task learning [2, 1], group lasso [16], and simultaneous sparse approximation [14, 15]) in the past years. In [1], the authors develop a convex framework for multi-task learning based on the $(2, 1)$-mixed norm formulation. Conditions for sparsity oracle inequalities and variable selection properties for a similar formulation are derived in [13], showing the advantage of joint estimation of tasks that share common support is statistically efficient. But the formulation has several drawbacks due to the looseness of its convex relaxation [12, 9]. The issue of bias that is inherent in the group lasso formulation was discussed in [12]. By defining a measure of sparsity level of the target signal under the group setting, the authors mention that the standard formulation of group lasso exhibits a bias that cannot be removed by simple reformulation of group lasso. In order to deal with this issue, recently [9] proposed the use of a non-convex regularizer and provided a numerical algorithm based on solving a sequence of convex relaxation problems. The method is based on a straightforward extension of a similar approach developed for the Lasso setting (cf., [17]), to the joint sparsity situation. Note that adaptive group-Lasso is a special case of [9]. In this paper, we propose a simple two-step procedure, to overcome the drawbacks of the standard group-Lasso relaxation. Compared to [9], the proposed approach is entirely convex and hence attains the global solution.

The current paper has two theoretical contributions. First, under a multi-task random effects model, we obtain an expected prediction error bound that relates the predictive performance to the accuracy of covariance estimation; by adapting high dimensional sparse covariance estimation procedures such as [8, 4], we can obtain consistent estimate of covariance matrix which leads to asymptotically optimal performance. Second, it is shown that under our random effects model, group Lasso in general does not accurately estimate the covariance matrix and thus is not optimal under the model considered. Experiments show that this approach provides improved performance compared to group Lasso (and the multi-stage versions) on simulated and real datasets.

## 2 Joint Sparsity Random Effects Model and Group Lasso

We consider joint sparsity regularization problems under multi-task learning. In multi-task learning, we consider $m$ linear regression problems tasks $\ell = 1, \ldots, m$

$$y^{(\ell)} = X^{(\ell)} \bar{\beta}^{(\ell)} + \epsilon^{(\ell)}. \quad (1)$$

We assume that each $y^{(\ell)}$ is an $n^{(\ell)}$ dimensional vector, each $X^{(\ell)}$ is an $n^{(\ell)} \times d$ dimensional matrix, each $\bar{\beta}^{(\ell)}$ is the target coefficient vector for task $\ell$ in $d$ dimension. For simplicity, we also assume that $\epsilon^{(\ell)}$ is an $n^{(\ell)}$ dimensional iid zero-mean Gaussian noise vector with variance $\sigma^2$: $\epsilon^{(\ell)} \sim N(0, \sigma^2 I_{n^{(\ell)} \times n^{(\ell)}})$.

The joint sparsity model in multi-task learning assumes that all $\bar{\beta}^{(\ell)}$ share similar supports: $\mathrm{supp}(\bar{\beta}^{(\ell)}) \subset \bar{F}$ for some common sparsity pattern $\bar{F}$, where $\mathrm{supp}(\beta) = \{j : \beta_j \neq 0\}$. The convex relaxation formulation for this model is given by group Lasso

$$\min_{\beta} \left[ \sum_{\ell=1}^{m} \frac{1}{2} \left\| y^{(\ell)} - X^{(\ell)} \beta^{(\ell)} \right\|_2^2 + \lambda \sum_{j=1}^{d} \sqrt{\sum_{\ell=1}^{m} (\beta_j^{(\ell)})^2} \right], \quad (2)$$

where $\beta = \{\beta^{(\ell)}\}_{\ell=1,\ldots,m}$.

We observe that the multi-task group Lasso formulation (2) is equivalent to $\min_{\beta,\omega} F(\beta, \omega)$, where $F(\beta, \omega) =$

$$\sum_{\ell=1}^{m} \frac{1}{2\sigma^2} \left\| y^{(\ell)} - X^{(\ell)} \beta^{(\ell)} \right\|_2^2 + \sum_{j=1}^{d} \frac{1}{2\omega_j} \sum_{\ell=1}^{m} (\beta_j^{(\ell)})^2$$

$$+ \frac{m}{2\sigma^2} \sum_{j=1}^{d} \omega_j \quad (3)$$

with $\lambda = \sigma \sqrt{m}$, where $\beta = \{\beta^{(\ell)}\}_{\ell=1,\ldots,m}$ and $\omega = \{\omega_j\}_{j=1,\ldots,d}$. With fixed hyper parameter $\omega$, we note that (2) is a special case of

$$\min_{\beta} \sum_{\ell=1}^{m} \frac{1}{2\sigma^2} \left\| y^{(\ell)} - X^{(\ell)} \beta^{(\ell)} \right\|_2^2 + \frac{1}{2} \sum_{\ell=1}^{m} (\beta^{(\ell)})^\top \Omega^{-1} \beta^{(\ell)}, \quad (4)$$

where $\Omega$ is a hyper parameter covariance matrix shared among the tasks. This general method employs a common quadratic regularizer that is shared by all the tasks. The group Lasso formulation (2) assumes a specific form of diagonal covariance matrix $\Omega = \mathrm{diag}(\{\omega_j\})$.

Equation (4) suggests the following random effects model for joint sparsity regularization, where the coefficient vectors $\bar{\beta}^{(\ell)}$ are random vectors generated independently for each task $\ell$; however they share the same covariance matrix $\bar{\Omega}$: $E \, \bar{\beta}^{(\ell)} \bar{\beta}^{(\ell)\top} = \bar{\Omega}$. Given the coefficient vector $\bar{\beta}$, we then generate $y^{(\ell)}$ based on (4). Note that we assume that $\Omega$ may contain zero-diagonal elements. If $\Omega_{jj} = 0$, then the corresponding $\bar{\beta}_j^{(\ell)} = 0$ for all $\ell$. Therefore we call this model *joint sparsity random effects model* for multi-task learning.

## 3 Joint Sparsity via Covariance Estimation

Under the proposed joint sparsity random effects model, it can be shown (see Section 4) that the optimal quadratic optimizer $(\beta^{(\ell)})^\top \Omega^{-1} \beta^{(\ell)}$ in (2) is obtained at the true covariance $\Omega = \bar{\Omega}$. This observation suggests the following estimation procedure involving two steps:

- Step 1: Estimate the joint covariance matrix $\Omega$ as hyper parameter. In particular, this paper suggests the following method as discussed in Section 3.1: $\hat{\Omega} =$

$$\arg\min_{\Omega \in \mathcal{S}} \left[ \frac{1}{2} \sum_{\ell=1}^{m} \left\| y^{(\ell)} y^{(\ell)\top} - X^{(\ell)} \Omega X^{(\ell)\top} \right\|_F^2 + R(\Omega) \right], \quad (5)$$

where $\|\cdot\|_F$ denotes the matrix Frobenius norm, $\mathcal{S}$ is the set of symmetric positive semi-definite matrices, and $R(\Omega)$ is an appropriately defined regularizer function (specified in Section 3.1).

- Step 2: Compute each $\beta^{(\ell)}$ separately given the estimated $\hat{\Omega}$ using:

$$\hat{\beta}^{(\ell)} = \left( X^{(\ell)\top} X^{(\ell)} + \lambda \hat{\Omega}^{-1} \right)^{-1} X^{(\ell)\top} y^{(\ell)}, \quad (6)$$

where $\ell = 1, \ldots, m$.

Note that the estimation method proposed in step 1 holds for a general class of covariance matrices. Meaningful estimates of the covariance matrix could be obtained even when the generative model assumption is violated. If the dimension $d$ and sample size $n$ per task are fixed, it can be shown relatively easily using classical asymptotic statistics that when $m \to \infty$, we can reliably estimate the true covariance $\bar{\Omega}$ using (5), i.e., $\hat{\Omega} \to \bar{\Omega}$. Therefore the method is asymptotically optimal as $m \to \infty$. On the other hand, the group Lasso formulation (3) produces sub-optimal estimate of $\omega_j$, as we shall see in Section 4.2. We would like to point out that in cases when the matrix $\hat{\Omega}$ is not invertible (for example, as in the sparse diagonal case as we see next) we replace the inverse with pseudo-inverse. For ease of presentation, we use the inverse throughout the presentation, though it should be clear from the context.

### 3.1 Sparse Covariance Coding Models

In our two step procedure, the covariance estimation of step 1 is more complex compared to step 2, which involves only the solutions of ridge regression problems. As mentioned above, if we employ a full covariance estimation model, then the estimation procedure proposed in this work is asymptotically optimal when $m \to \infty$. However, since modern asymptotics are often concerned with the scenario when $d \gg n$, computing a $d \times d$ full matrix $\Omega$ becomes impossible without further structure on $\Omega$. In this section, we assume that $\Omega$ is diagonal, which is consistent with the group Lasso model.

This section explains how to estimate $\Omega$ using our generative model, which implies that $\bar{\beta}^{(\ell)} \sim N(0, \Omega)$, and $y^{(\ell)} = X^{(\ell)} \bar{\beta}^{(\ell)} + \epsilon^{(\ell)}$ with $\epsilon^{(\ell)} \sim N(0, \sigma^2 I_{n^{(\ell)} \times n^{(\ell)}})$. Taking expectation of $y^{(\ell)} y^{(\ell)\top}$ with respect to $\epsilon$ and $\bar{\beta}^{(\ell)}$, we obtain $E_{\bar{\beta}^{(\ell)}, \epsilon} y^{(\ell)} y^{(\ell)\top} = X^{(\ell)} \Omega X^{(\ell)\top} + \sigma^2 I_{n^{(\ell)} \times n^{(\ell)}}$. This suggests the following estimator of $\Omega$: $\hat{\Omega} =$

$$\arg\min_{\Omega \in \mathcal{S}} \sum_{\ell=1}^{m} \left\| y^{(\ell)} y^{(\ell)\top} - X^{(\ell)} \Omega X^{(\ell)\top} - \sigma^2 I_{n^{(\ell)} \times n^{(\ell)}} \right\|_F^2,$$

where $\|\cdot\|_F$ is the matrix Frobenius norm. This is equivalent to

$$\hat{\Omega} = \arg\min_{\Omega \in \mathcal{S}} \frac{1}{2} \sum_{\ell=1}^{m} \left\| y^{(\ell)} y^{(\ell)\top} - X^{(\ell)} \Omega X^{(\ell)\top} \right\|_F^2$$
$$+ \lambda \mathrm{tr}\left( \Omega \sum_{\ell=1}^{m} X^{(\ell)\top} X^{(\ell)} \right) \quad (7)$$

with $\lambda = \sigma^2$. Similar ideas for estimating covariance by this approach appeared in [8, 5]. We may treat the last term as regularizer of $\Omega$, and in such sense a more general form is to consider $\hat{\Omega} =$

$$\arg\min_{\Omega \in \mathcal{S}} \left[ \frac{1}{2} \sum_{\ell=1}^{m} \left\| y^{(\ell)} y^{(\ell)\top} - X^{(\ell)} \Omega X^{(\ell)\top} \right\|_F^2 + R(\Omega) \right],$$

where $R(\Omega)$ is a general regularizer function of $\Omega$. Note that the dimension $d$ can be large, and thus special structure is needed to regularize $\Omega$. In particular, to be consistent with group Lasso, we impose the diagonal covariance constraint $\Omega = \mathrm{diag}(\{\omega_j\})$, and then encourage sparsity as follows: $\hat{\Omega} =$

$$\arg\min_{\{\omega_j \geq 0\}} \sum_{\ell=1}^{m} \frac{1}{2} \| y^{(\ell)} y^{(\ell)\top} - X^{(\ell)} \mathrm{diag}(\{\omega_j\}) X^{(\ell)\top} \|_F^2$$
$$+ \lambda \sum_j \omega_j. \quad (8)$$

This formulation leads to sparse estimation of $\omega_j$, which we call *sparse covariance coding (scc)*. Note that the above optimization problem is convex and hence the solution could be computed efficiently. This formulation is consistent with the group Lasso regularization which also assumes diagonal covariance implicitly as in (2). It should be noted that if the diagonals of $\sum_{\ell=1}^{m} X^{(\ell)\top} X^{(\ell)}$ have identical values, then up to a rescaling of $\lambda$, (8) is equivalent to (7) with $\Omega$ restricted to be a diagonal matrix. In the experiments conducted on real world data sets, there was no significant difference between the two regularization terms (see Table 4), when both formulations are restricted to diagonal $\Omega$.

### 3.2 Other Covariance Coding Models

We now demonstrate the generality of the proposed approach for multi-task learning. Note that in addition to the sparse covariance coding method (8) that assumes a diagonal form of $\Omega$ plus sparsity constraint, some other structures may be explored. One method that has been suggested for covariance estimation in [4] is the following formulation:

$$\hat{\Omega} = \arg\min_{\Omega \in \mathcal{S}} \sum_{\ell=1}^{m} \|y^{(\ell)} y^{(\ell)\top} - X^{(\ell)} \Omega X^{(\ell)}\|_F^2$$
$$+ 2\lambda \sum_k \gamma_k \sqrt{\sum_m \Omega_{k,m}^2}, \quad (9)$$

where $\mathcal{S}$ denotes the set of symmetric positive semi-definite matrices $\mathcal{S}$. This approach selects a set of features, and then models a full covariance matrix within the selected set of features. Although the feature selection is achieved with a group Lasso penalty, unlike this work, [4] didn't study the possibility of using covariance estimation to do joint feature selection (which is the main purpose of this work), but rather studied covariance estimation as a separate problem.

The partial full covariance model in (9) has complexity in between that of the full covariance model and the sparse diagonal covariance model (sparse covariance coding) which we promote in this paper, at least for the purpose of joint feature selection. The latter has the smallest complexity, and thus more effective for high dimensional problems that tend to cause over-fitting.

Another model with complexity in between of sparse diagonal covariance and full covariance model is to model the covariance matrix $\Omega$ as the sum of a sparse diagonal component plus a low-rank component. This is similar in spirit to the more general sparse+low-rank matrix decomposition formulation recently appeared in the literature [7, 6, 11]. However since the sparse matrix is diagonal, identifiability holds trivially (as described in the appendix) and hence one could in principal, recover both the diagonal and the low-rank objects individually which preserves the advantages of the diagonal formulation and the richness of low-rank formulation. The model assumption is $\Omega = \Omega_S + \Omega_L$, where $\Omega_S$ is the diagonal matrix and $\Omega_L$ is the low-rank matrix. The estimation procedure now becomes the following optimization problem (and the rest follows) $[\hat{\Omega}_S, \hat{\Omega}_L] =$

$$\arg\min_{\Omega_S, \Omega_L} \sum_{\ell=1}^{m} \frac{1}{2} \|y^{(\ell)} y^{(\ell)\top} - X^{(\ell)}(\Omega_S + \Omega_L) X^{(\ell)\top}\|_F^2$$
$$+ \lambda_1 \|\Omega_S\|_{\text{vec}(1)} + \lambda_2 \|\Omega_L\|_*,$$

subject to the condition that $\Omega_S$ is a non-negative diagonal matrix, and $\Omega_L \in \mathcal{S}$, where $\|\cdot\|_{\text{vec}(1)}$ is the element-wise $L1$ norm and $\|\cdot\|_*$ corresponds to trace-norm.

## 4 Theoretical Analysis

In this section we do a theoretical analysis of the proposed method. Specifically, we first derive upper and lower bounds for prediction error for the joint sparsity random effects model and show the optimality of the proposed approach. Informally, the notion of optimality considered is as follows: what is the 'optimal shared quadratic regularizer', when $m$ and $d$ goes to infinity and when solutions for each task can be written as individual ridge regression solutions with a shared quadratic regularizer (note that this includes group-Lasso method). Next, we demonstrate with a simple example (i.e., considering the low-dimensional setting) the drawback of the standard group-Lasso relaxation. In a way, this example also serves as a motivation for the approach proposed in this work and provides concrete intuition.

We consider a simplified analysis with $\hat{\Omega}$ replaced by $\hat{\Omega}^{(\ell)}$ in Step 2 so that $\hat{\Omega}^{(\ell)}$ does not depend on $y^{(\ell)}$:

$$\hat{\beta}^{(\ell)} = \left(X^{(\ell)\top} X^{(\ell)} + \lambda \hat{\Omega}^{(\ell)-1}\right)^{-1} X^{(\ell)\top} y^{(\ell)}. \quad (10)$$

For example, this can be achieved by replacing Step 1 with $\hat{\Omega}^{(\ell)} =$

$$\arg\min_{\Omega \in \mathcal{S}} \left[\frac{1}{2} \sum_{k \neq \ell} \left\|y^{(k)} y^{(k)\top} - X^{(k)} \Omega X^{(k)\top}\right\|_F^2 + R(\Omega)\right]. \quad (11)$$

Obviously when $m$ is large, we have $\hat{\Omega}^{(\ell)} \approx \hat{\Omega}$. Therefore the analysis can be slightly modified to the original formulation, with an extra error term of $O(1/m)$ that vanishes when $m \to \infty$. Nevertheless, the independence of $\hat{\Omega}^{(\ell)}$ and $y^{(\ell)}$ simplifies the argument and makes the essence of our analysis much easier to understand.

### 4.1 Prediction Error

This section derives an expected prediction error bound for the coefficient vector $\hat{\beta}^{(\ell)}$ in (10) in terms of the accuracy of the covariance matrix estimation $\hat{\Omega}^{(\ell)}$. We consider the fixed design scenario, where the design matrices $X^{(\ell)}$ are fixed and $\epsilon^{(\ell)}$ and $\bar{\beta}^{(\ell)}$ are random.

**Theorem 4.1.** *Assume that $\lambda \geq \sigma^2$. For each task $\ell$, given $\hat{\Omega}^{(\ell)}$ that is independent of $y^{(\ell)}$, the expected prediction error with $\hat{\beta}^{(\ell)}$ in (10) is bounded as*

$$\sigma^2 \lambda \omega^{(\ell)} \leq A \leq \lambda^2 \omega^{(\ell)},$$

*where* $A = E \|X^{(\ell)} \hat{\beta}^{(\ell)} - X^{(\ell)} \bar{\beta}^{(\ell)}\|_2^2 - \left\|X^{(\ell)} \bar{\Omega}^{1/2} \left(\bar{\Omega}^{1/2} \Sigma^{(\ell)} \bar{\Omega}^{1/2} + \lambda I\right)^{-1/2}\right\|_F^2$ *and the expectation is with respect to the random effects $\bar{\beta}^{(\ell)}$ and*

noise $\epsilon^{(\ell)}$, and $\Sigma^{(\ell)} = X^{(\ell)\top} X^{(\ell)}$, and

$$\omega^{(\ell)} = \|X^{(\ell)} \left(\hat{\Omega}^{(\ell)}\Sigma^{(\ell)} + \lambda I\right)^{-1} (\hat{\Omega}^{(\ell)} - \bar{\Omega})(\Sigma^{(\ell)})^{1/2}$$
$$\left((\Sigma^{(\ell)})^{1/2}\bar{\Omega}(\Sigma^{(\ell)})^{1/2} + \lambda I\right)^{-1/2} \|_F^2.$$

The bound shows that the prediction performance of (10) depends on the accuracy of estimating $\bar{\Omega}$. In particular, if $\hat{\Omega}^{(\ell)} = \bar{\Omega}$, then the optimal prediction error of $\left\|X^{(\ell)}\bar{\Omega}^{1/2}\left(\bar{\Omega}^{1/2}X^{(\ell)\top}X^{(\ell)}\bar{\Omega}^{1/2} + \lambda I\right)^{-1/2}\right\|_F^2$ can be achieved. A simplified upper bound is $E \, \|X^{(\ell)}\hat{\beta}^{(\ell)} - X^{(\ell)}\bar{\beta}^{(\ell)}\|_2^2 \leq \left\|X^{(\ell)}\bar{\Omega}^{1/2}\left(\bar{\Omega}^{1/2}\Sigma^{(\ell)}\bar{\Omega}^{1/2} + \lambda I\right)^{-\frac{1}{2}}\right\|_F^2 + \lambda^{-1}\|\Sigma^{(\ell)}(\hat{\Omega}^{(\ell)} - \bar{\Omega})\|_F^2$.

This means that if the covariance estimation is consistent; that is, if $\hat{\Omega}^{(\ell)}$ converges to $\bar{\Omega}$, then our method achieves the optimal prediction error $\left\|X^{(\ell)}\bar{\Omega}^{1/2}\left(\bar{\Omega}^{1/2}\Sigma^{(\ell)}\bar{\Omega}^{1/2} + \lambda I\right)^{-1/2}\right\|_F^2$ for all tasks.

The consistency of $\hat{\Omega}^{(\ell)}$ has been studied in the literature, for example by [4] under high dimensional sparsity assumptions. Such results can be immediately applied with Theorem 4.1 to obtain optimality of the proposed approach. Specifically, we consider the case of diagonal covariance matrix, where the sparsity in $\bar{\Omega}$ is defined as the number of non-zero diagonal entries, i.e., $s = |\{i : \Omega_{ii} \neq 0\}|$. Following [4], we consider the case $X^{(\ell)} = X \in \mathbb{R}^{n \times d}, \ell = 1, \ldots, m$. Let $X_J$ denote the sub matrix of $X$ obtained by removing the columns of $X$ whose indices are not in the set $J$. We also assume that the diagonals of $X^\top X$ have identical values so that (8) is equivalent to (7) up to a scaling of $\lambda$.

Let $\rho_{\min}(A)$ and $\rho_{\max}(A)$ for a matrix $A$ denote the smallest and largest eigenvalue of $A$ respectively. We introduce two quantities [4] that impose certain assumptions on the matrix $X$.

**Definition 1.** *For $0 < t \leq d$, define $\rho_{\min}(t) := \inf_{\substack{J \subset \{1,\ldots,d\} \\ |J| \leq t}} \rho_{\min}(X_J^\top X_J)$.*

**Definition 2.** *The mutual coherence of the columns $X_t, t = 1, \ldots, d$ of $X$ is defined as $\theta(X) := \max\{|X_{t'}^\top X_t|, t \neq s', 1 \leq t, t' \leq d\}$ and let $X_{\max}^2 := \max\{\|X_t\|_2^2, 1 \leq t \leq d\}$.*

We now state the following theorem establishing the consistency of covariance estimation (given by Eq 11) in the high-dimensional setting. The proof essentially follows the same argument for Theorem 8 in [4], by noticing the equivalence between (8) and (7), which implies consistency.

**Theorem 4.2.** *Assume that $\bar{\Omega}$ is diagonal, and $\theta(X) < \rho_{\min}(s)^2/4\rho_{\max}(X^\top X)s$. Assume $n$ is fixed and the number of tasks and dimensionality $m, d \to \infty$ such that $\sqrt{s} \ln d/m \to 0$. Then the covariance estimator of (11), with appropriately chosen $\lambda$ and $R(\Omega)$ defined by (8), converges to $\bar{\Omega}$:*

$$\|X(\hat{\Omega}^{(\ell)} - \bar{\Omega})X^\top\|_F^2 \to_P 0. \quad (12)$$

The following corollary, which is an immediate consequence of Theorem 4.1 and 4.2, establishes the asymptotic optimality (for prediction) of the proposed approach under the sparse diagonal matrix setting and $R(\Omega)$ defined as in (8). Similar result could be derived for other regularizers for $R(\Omega)$.

**Corollary 1.** *Under the assumption of Theorem 4.1 and 4.2, the two-step approach defined by (11) and (10), with $R(\Omega)$ defined by (8) is asymptotically optimal for prediction, for each task $\ell$:*

$$E \, \|X\hat{\beta}^{(\ell)} - X\bar{\beta}^{(\ell)}\|_2^2$$
$$- \left\|X\bar{\Omega}^{1/2}\left(\bar{\Omega}^{1/2}X^\top X\bar{\Omega}^{1/2} + \lambda I\right)^{-1/2}\right\|_F^2 \to_P 0.$$

Note that the asymptotics considered above, reveals the advantage of *multi-task learning* under the joint sparsity assumption: with a fixed number of samples per each task, as the dimensions of the samples and *number of tasks* tend to infinity (obeying the condition given in theorem 4.2) the proposed two-step procedure is asymptotically optimal for prediction. Although for simplicity, we state the optimality result for (11) and (10), the same result holds for the two-step procedure given by (5) and (6), because $\hat{\Omega}^{(\ell)}$ of (11) and $\hat{\Omega}$ of (5) differ only by a factor of $O(1/m)$ which converges to zero under the asymptotics considered. Finally, we would like to remark that the mutual coherence assumption made in Theorem 4.2 could be relaxed to milder conditions (based on restricted eigenvalue type assumptions) - we leave it as future work.

### 4.2 Drawback of Group Lasso

In general, group Lasso does not lead to optimal performance due to looseness of the single step convex relaxation. [12, 9]. This section presents a simple but concrete example to illustrate the phenomenon and shows how $\bar{\Omega}$ is under-estimated in the group-Lasso formulation. Combined with the previous section, we have a complete theoretical justification of the superiority of our approach over group Lasso, which we will also demonstrate in the empirical study.

For this purpose, we only need to consider the following relatively simple illustration (in the low-dimensional setting). We consider the case when all design matrices equal identity: $X^{(\ell)} = I$ for $\ell = 1, \ldots, m$. This formulation is similar to *Normal means models*, a popular model in the statistics literature. It is instructive to consider this model because of its closed form solution. It helps in deriving useful insights that further help for a better understanding of more general cases. We are interested in the asymptotic behavior when $m \to \infty$ (with $n^{(\ell)}$ and $d$ fixed), which simplifies the analysis, but nevertheless reveals the problems associated with the standard group Lasso formulation. Moreover, it should be mentioned that although the two-step procedure is motivated from a generative model, the analysis presented in this section does not need to assume that each $\beta^{(\ell)}$ is truly generated from such a model.

**Proposition 1.** *Suppose that $n^{(\ell)} = d$ and $X^{(\ell)} = I$ for $\ell = 1, \ldots, m$, and $m \to \infty$. The sparse covariance estimate corresponding to the formulation defined by (8) is consistent.*

*Proof.* The sparse covariance coding formulation (8) is equivalent to (with the intention of setting $\lambda = \sigma^2$): $\hat{\Omega}^{scc} = \arg\min_{\{\omega_j \geq 0\}} \sum_{\ell=1}^{m} \frac{1}{2} \left\| y^{(\ell)} y^{(\ell)\top} - \text{diag}(\{\omega_j\}) \right\|_F^2 + \lambda m \sum_j \omega_j$. The closed form solution is given by $\hat{\omega}_j^{scc} = \max\left(0, m^{-1} \sum_{\ell=1}^{m} (y_j^{(\ell)})^2 - \lambda\right)$ for $j = 1, \ldots, d$. Since $m^{-1} \sum_{\ell=1}^{m} (y_j^{(\ell)})^2 \to E_{\beta^{(\ell)}}(\beta_j^{(\ell)})^2 + \sigma^2$ as $m \to \infty$, the variance $\hat{\omega}_j^{scc} \to E_{\beta^{(\ell)}}(\beta_j^{(\ell)})^2$ with $\lambda = \sigma^2$. Therefore $\hat{\omega}_j$ is consistent. □

Note that by plugging-in the estimate of variance into (6) with the same $\lambda$ (with $\lambda = \sigma^2$), we obtain

$$\hat{\beta}_j^{(\ell)} = y_j^{(\ell)} \max\left(0, 1 - \frac{\lambda}{m^{-1} \sum_{\ell=1}^{m} (y_j^{(\ell)})^2}\right). \quad (13)$$

An immediate consequence of Proposition 1 is that the estimate define in (13) is asymptotically optimal for any method using a quadratic regularizer shared by all the tasks.

A similar analysis of group Lasso formulation would reveal its drawback. Consider the group Lasso formulation defined in (3). Under similar settings, the formulation can be written as $[\hat{\beta}, \hat{\omega}^{gl}] =$

$$\arg\min_{\beta, \omega} \sum_{\ell=1}^{m} \left\| y^{(\ell)} - \beta^{(\ell)} \right\|_2^2 + \lambda \sum_{j=1}^{d} \frac{1}{\omega_j} \sum_{\ell=1}^{m} (\beta_j^{(\ell)})^2$$

$$+ m \sum_{j=1}^{d} \omega_j.$$

The closed form solution for the above formulation is given by $\hat{\omega}_j^{gl} = \max\left(0, \sqrt{\lambda m^{-1} \sum_{\ell=1}^{m}(y_j^{(\ell)})^2} - \lambda\right)$, for $j = 1, \ldots, d$, and the corresponding coefficient estimate is $\hat{\beta}_j^{(\ell)} = y_j^{(\ell)} \max\left(0, 1 - \frac{\sqrt{\lambda}}{\sqrt{m^{-1} \sum_{\ell=1}^{m}(y_j^{(\ell)})^2}}\right)$, for $\ell = 1, \ldots, m$ and $j = 1, \ldots, d$.

The solution for $\hat{\omega}_j^{gl}$ implies that it is not possible to pick a fixed $\lambda$ such that the group Lasso formulation gives consistent estimate of $\omega_j$. Since from (3), it is evident that group Lasso can also be regarded as a method that uses a quadratic regularizer shared by all the tasks, we know that the solution obtained for the corresponding co-efficient estimate is asymptotically sub-optimal. In fact, the covariance estimate $\hat{\omega}_j^{gl}$ is significantly smaller than the correct estimate $\hat{\omega}_j^{scc}$. This under-estimate of $\omega_j$ in group Lasso implies a corresponding under-estimate of $\beta^{(\ell)}$ obtained via group Lasso, when compared to (13). This under-estimation is the underlying theoretical reason why the proposed two-step procedure is superior to group Lasso for learning with joint sparsity. This claim is also confirmed by our empirical studies.

## 5 Experiments

We demonstrate the advantage of the proposed two-step procedure through (i) multi-task learning experiments on synthetic and real-world data sets and (ii) sparse covariance coding based image classification.

### 5.1 Multi-task learning

We first report illustrative experiments conducted on synthetic data sets with the proposed models. They are compared with the standard group-lasso formulation. The experimental set up is as follows: the number of tasks $m = 30$, $d = 256$, and $n^\ell = 150$. The data matrix consists of entries from standard Gaussian $N(0, 1)$. To generate the sparse co-efficients, we first generate a random Gaussian vector in $d$ dimensions and set to zero $d - k$ of the co-efficients to account for sparsity. The cardinality of the set of non-zero coefficients is varied as $k = 50, 70, 90$ and the noise variance was $0.1$. The results reported are averages over $100$ random runs. We compare against standard group lasso, MSMTFL [9] (note that this is a non-convex approach, solved by sequence of convex relaxations) and another natural procedure (GLS-LS) where one uses group lasso for feature selection and with the selected features, one does least squares regression to estimate the coefficients. A precise theoretical comparison to MSMTFL procedure is left as future work.

Tables 2 shows the coefficient estimation error when the samples are such that they share $80\%$ as common basis (and the rest $20\%$ is selected randomly from the remaining basis) and when the samples share the same indices of non-zero coefficients (and the actual values vary for each signals). We note that in both cases, the model with diagonal covariance assumption and partial full covariance (Equation 9) outperforms the standard group lasso formulation, with the diagonal assumption performing better because of good estimates. The diagonal+low-rank formulation slightly outperforms the other models as it preserves the advantages of the diagonal model, while at the same time allows for additional modeling capability through the low-rank part, through proper selection of regularization parameters by cross-validation.

**Support selection:** While the above experiment sheds light on co-efficient estimation error, we performed another experiment to examine the selection properties of the proposed approach. Table 1 shows the hamming distance between selected basis and the actual basis using the different models. Note that Hamming distance is a desired metric for practical applications where exact recovery of the support set is not possible due to low signal-to-noise ratio. The indices with non-zero entry along the diagonal in the model with diagonal covariance assumption correspond to the selected basis. Similarly, indices with non-zero columns (or rows by symmetry) correspond to the selected basis in the partial full covariance model. The advantage of the diagonal assumption for joint feature selection is clearly seen from the table. This superiority in the feature selection process also explains the better performance achieved for coefficient estimation. A rigorous theoretical study of the feature selection properties is left as future work.

**Correlated data:** We next study the effect of correlated data set on the proposed approach. We generated correlated Gaussian random variables (corresponding to the size of the data matrix) in order to fill the matrix $X$ for each task. The correlation co-efficient was fixed at $0.5$. We worked with fully overlapped support set. Other problem parameters were retained. We compared the estimation accuracy of the proposed approach with different settings with group lasso and its variants. The results are summarized in Table 3. Note that the proposed approach performs much better than the group-Lasso based counterparts. Precisely characterizing this improvement theoretically would be interesting.

Next, the proposed approach was tested on three standard multi-task regression datasets (computer, school and sarcos datasets) and compared with the standard approach for multi-task learning: mixed $(2, 1)$-norms or group lasso (2). A description of the datasets is given below:

**Computer data set:** This dataset consists of a survey among 180 people (corresponding to tasks). Each rated the likelihood of purchasing one of 20 different computers. The input consists 13 different computer characteristics, while the output corresponds to ratings. Following [1], we used the first 8 examples per task for training and the last 4 examples per task for testing.

**School data set:** This dataset is from the London Education Authority and consists of the exam scores of 15362 students from 139 schools (corresponding to tasks). The input consists 4 school-based and 3 student-based attributes, along with the year. The categorical features are replaced with binary features. We use 75% of the data set for training and the rest for testing.

**Sarcos data set:** The dataset[1] has 44,484 train samples and 4449 test samples. The task is to map a 21-dimensional input space (corresponding to characteristics of robotic arm) to the the output corresponding to seven torque measurement (tasks) to predict the inverse dynamics.

We report the average (accross tasks) root mean square error on the test data set in Table 4. Note that the proposed two-step approach performs better than the group lasso approach on all the data sets. The data sets correspond to cases with varied data size and number of tasks. Observe that even with a small training data (computer data set), performance of both our approach is better than the group-lasso approach.

### 5.2 SCC based Image Classification

In this section, we present a novel application of the proposed approach for obtaining sparse codes for gender recognition in CMU Multi-pie data set. The database contains 337 subjects (235 male and 102 female) across simultaneous variations in pose, expression, and illumination. The advantages of jointly coding the extracted local descriptors of an image with respect to a given dictionary for the purpose of classification has been highlighted in [3]. They propose a method based on mixed $(2, 1)$-norm to jointly find a sparse representation of an image based on local descriptors of that image. Following a similar experimental setup, we use the proposed sparse covariance coding approach for attaining the same goal.

Each image is of size $30 \times 40$, size of patches is $8 \times 8$, and number of overlapping patches per image is 64. Local descriptors for each images are extracted in the form of overlapping patches and a dictionary is learned based on the obtained patches by sparse coding. With the learnt dictionary, the local descriptors of each image is jointly sparse coded via the diagonal covariance matrix assumption and the codes thus obtained are used or classification. This approach is compared with the group sparse coding based approach. Linear SVM is used in the final step for classification. Note that the purpose of the experiment is not learning a dictionary. Table 5 shows the test set and train set error for the classifier thus obtained. Note that the proposed sparse covariance coding based approach outperforms the group sparse coding based approach for gender classification due to its better quality estimates.

|  | Group sparse coding | Sparse cov. coding |
|---|---|---|
| Train error | $6.67 \pm 1.34\%$ | $5.56 \pm 1.62\%$ |
| Test error | $7.48 \pm 1.54\%$ | $6.32 \pm 1.12\%$ |

Table 5: Face image classification based on gender: Test and Train set error rates for sparse covariance coding and group sparse coding (both with a fixed dictionary).

---
[1] http://www.gaussianprocess.org/gpml/data/

| Method | 80% shared basis | | | Completely shared basis | | |
|---|---|---|---|---|---|---|
| | k=50 | k=70 | k=90 | k=50 | k=70 | k=90 |
| Standard group lasso | 0.18 | 0.22 | 0.27 | 0.11 | 0.16 | 0.22 |
| MSMTFL | 0.15 | 0.18 | 0.20 | 0.07 | 0.08 | 0.17 |
| Partial full covariance | 0.17 | 0.20 | 0.23 | 0.07 | 0.11 | 0.16 |
| Sparse diagonal covariance | 0.13 | 0.16 | 0.20 | 0.05 | 0.09 | 0.14 |

Table 1: Support selection: Hamming distance between true non-zero indices and estimated non-zero indices by the indicated method for all signals.

| Method | k=50 | k=70 | k=90 |
|---|---|---|---|
| standard group Lasso | $0.1541 \pm 0.0045$ | $0.1919 \pm 0.0092$ | $0.2404 \pm 0.0124$ |
| GLS-LS | $0.1498 \pm 0.0032$ | $0.1901 \pm 0.0034$ | $0.2383 \pm 0.0342$ |
| Partial full covariance | $0.1239 \pm 0.0063$ | $0.1542 \pm 0.0131$ | $0.1992 \pm 0.0143$ |
| Sparse Diagonal covariance | $0.1022 \pm 0.0054$ | $0.1393 \pm 0.0088$ | $0.1701 \pm 0.0104$ |
| MSMTFL | $0.1276 \pm 0.0075$ | $0.1564 \pm 0.0153$ | $0.1987 \pm 0.0201$ |
| Diag+Low-rank covariance | $0.1031 \pm 0.0042$ | $0.1212 \pm 0.0122$ | $0.1532 \pm 0.0173$ |
| | | | |
| Standard group Lasso | $0.1032 \pm 0.0086$ | $0.1574 \pm 0.0151$ | $0.1733 \pm 0.0190$ |
| GLS-LS | $0.1010 \pm 0.0045$ | $0.1532 \pm 0.0134$ | $0.1698 \pm 0.0430$ |
| Partial full covariance | $0.0735 \pm 0.0078$ | $0.1131 \pm 0.0148$ | $0.1576 \pm 0.0201$ |
| Sparse Diagonal covariance | $0.0447 \pm 0.0071$ | $0.0828 \pm 0.0165$ | $0.1184 \pm 0.0198$ |
| MSMTFL | $0.0643 \pm 0.0093$ | $0.0832 \pm 0.0200$ | $0.1457 \pm 0.0223$ |
| Diag+low-rank Covariance | $0.0452 \pm 0.0084$ | $0.0786 \pm 0.0136$ | $0.1012 \pm 0.0161$ |

Table 2: Coefficient estimation: Normalized $L_2$ distance between true coefficients and estimated coefficients by the indicated method. First 5 rows correspond to 80% shared basis and the last 5 rows correspond to fully shared basis.

## 6 Discussion and Future work

We proposed a two-step estimation procedure based on a specialized random effects model for dealing with joint sparsity regularization and demonstrated its advantage over the group-Lasso formulation. The proposed approach highlights the fact that enforcing interesting structure on covariance of the coefficients is better for obtaining joint sparsity in the coefficients. We leave a theoretical comparison to the MSMTFL procedure, precisely quantifying the statistical improvement provided by the proposed approach (note that MSMTFL being a non-convex procedure does not attain the global optimal solution [9]) as future work. Future work also includes (i) relaxing the assumptions made in the theoretical analysis, (ii) exploring more complex models like imposing group-mean structure on the parameters for additional flexibility, (iii) other additive decomposition of the covariance matrix with complementary regularizers and (iv) using locally-smoothed covariance estimates for time-varying joint sparsity.

## A Identifiability of additive structure

The issue of identifiability (which is necessary subsequently for consistency and recovery guarantees) arises when we deal with additive decomposition of the covariance matrix. Here, we discuss about the conditions under which the model is identifiable, i.e., there exist an unique decomposition of the covariance matrix as the summation of the sparse diagonal matrix and low-rank matrix. We follow the discussion used in [11]. Let $\Omega = \Omega_s + \Omega_L$ denote the decomposition where $\Omega_s$ denotes the sparse diagonal matrix and $\Omega_L$ a low-rank matrix. Intuitively, identifiability holds if the sparse matrix is not low-rank (i.e., the support is sufficiently spread out) and the low-rank matrix is not too sparse (i.e., the singular vectors are away from co-ordinate axis). A formal argument is made based on the above intuition. We defined the following quantities (following [11]) below that measures the non-zero entries in any row or column of $\Omega_s$ and sparseness of the singular vectors of $\Omega_L$:

$$\alpha = \max\{\|\text{sign}(\Omega_s)\|_{1\to 1}, \|\text{sign}(\Omega_s)\|_{\infty\to\infty}\}$$

and

$$\beta = \|UU^T\|_\infty + \|VV^T\|_\infty + \|U\|_{2\to\infty}\|V\|_{2\to\infty},$$

where $U, V \in \mathbb{R}^{d\times r}$ are the left and right orthonormal singular vectors corresponding to non-zero singular values of $\Omega_L$ and $\|M\|_{p\to q} \stackrel{\text{def}}{=} \{\|Mv\|_q : v \in \mathcal{R}^m, \|v\|_p \leq 1\}$.

Note that, for a diagonal matrix, $\|\text{sign}(\Omega_s)\|_{1\to 1} = \|\text{sign}(\Omega_s)\|_{\infty\to\infty} = 1$. It is proved in [11] that if $\alpha\beta < 1$, then the matrices are identifiable, i.e, the sparse plus low-rank decomposition is unique. Therefore we only need to require $\beta < 1$ for identifiability, which is a rather weak assumption, satisfied by most low-rank matrices with sufficient spread of the support.

| Method | k=50 | k=70 | k=90 |
|---|---|---|---|
| Group Lasso | $0.2012 \pm 0.0033$ | $0.2655 \pm 0.0132$ | $0.3252 \pm 0.0323$ |
| GLS-LS | $0.2090 \pm 0.0098$ | $0.2702 \pm 0.0042$ | $0.3304 \pm 0.0333$ |
| Partial full covariance | $0.1706 \pm 0.0064$ | $0.2376 \pm 0.0224$ | $0.2701 \pm 0.0323$ |
| Sparse diagonal covariance | $0.1634 \pm 0.0022$ | $0.2112 \pm 0.0073$ | $0.2601 \pm 0.0231$ |
| MSMTFL | $0.1786 \pm 0.0023$ | $0.2323 \pm 0.0434$ | $0.2776 \pm 0.0223$ |
| Diag+Low-rank covariance | $0.1531 \pm 0.0042$ | $0.2002 \pm 0.0236$ | $0.2544 \pm 0.0145$ |

Table 3: Coefficient estimation: Normalized $L_2$ distance between true coefficients and estimated coefficients by the indicated method with correlated input data.

| Data set | Group lasso | MSMTFL | Sparse diagonal Covariances | Corr. Sparse diag (Eq.7) |
|---|---|---|---|---|
| Computer | $1.542 \pm 0.043$ | $1.334 \pm 0.031$ | $1.223 \pm 0.033$ | $1.209 \pm 0.054$ |
| School | $2.202 \pm 0.038$ | $2.033 \pm 0.241$ | $1.987 \pm 0.040$ | $2.012 \pm 0.073$ |
| Sarcos | $9.221 \pm 0.051$ | $9.113 \pm 0.145$ | $8.983 \pm 0.043$ | $9.002 \pm 0.032$ |

Table 4: Multi-task learning: Average (across task) MSE error on the test data set.

## B  Proof of Theorem 4.1

For notational simplicity, we remove the superscripts $(\ell)$ in the following derivation (e.g., denote $X^{(\ell)}$ by $X$, $\hat{\beta}^{(\ell)}$ by $\hat{\beta}$ and so on). We have the following decomposition

$$E\|X\hat{\beta} - X\bar{\beta}\|_2^2$$
$$= E\|X\big((X^\top X + \lambda\hat{\Omega}^{-1})^{-1}X^\top(X\bar{\beta} + \epsilon) - \bar{\beta}\big)\|_2^2$$
$$= E\|X(X^\top X + \lambda\hat{\Omega}^{-1})^{-1}\lambda\hat{\Omega}^{-1}\bar{\beta}\|_2^2$$
$$\quad + E\|X(X^\top X + \lambda\hat{\Omega}^{-1})^{-1}X^\top\epsilon\|_2^2$$
$$= \lambda^2 \mathrm{tr}\big[X(X^\top X + \lambda\hat{\Omega}^{-1})^{-1}\hat{\Omega}^{-1}\bar{\Omega}\hat{\Omega}^{-1}(X^\top X + \lambda\hat{\Omega}^{-1})^{-1}X^\top\big]$$
$$\quad + \sigma^2 \mathrm{tr}\big[X(X^\top X + \lambda\hat{\Omega}^{-1})^{-1}$$
$$\quad X^\top X(X^\top X + \lambda\hat{\Omega}^{-1})^{-1}X^\top\big]$$
$$\le \mathrm{tr}\lambda\big[X(\hat{\Omega}X^\top X + \lambda I)^{-1}(\lambda\bar{\Omega} + \hat{\Omega}X^\top X\hat{\Omega})$$
$$\quad (X^\top X\hat{\Omega} + \lambda I)^{-1}X^\top\big]$$
$$= \lambda(A + B + C)$$

where with $\Delta\hat{\Omega} = \hat{\Omega} - \bar{\Omega}$, we have $A = \mathrm{tr}\big[X(\hat{\Omega}X^\top X + \lambda I)^{-1}\Delta\hat{\Omega}X^\top X\Delta\hat{\Omega}(X^\top X\hat{\Omega} + \lambda I)^{-1}X^\top\big]$ and $B = 2\mathrm{tr}\big[X(\hat{\Omega}X^\top X + \lambda I)^{-1}\bar{\Omega}X^\top X\Delta\hat{\Omega}(X^\top X\hat{\Omega} + \lambda I)^{-1}X^\top\big]$ and $C = \mathrm{tr}\big[X(\hat{\Omega}X^\top X + \lambda I)^{-1}(\bar{\Omega}X^\top X\bar{\Omega} + \lambda\bar{\Omega})(X^\top X\hat{\Omega} + \lambda I)^{-1}X^\top\big]$. We can further expand $C$ as

$$C = \mathrm{tr}\big[X(\bar{\Omega}X^\top X + \lambda I)^{-1}(\bar{\Omega}X^\top X\bar{\Omega} + \lambda\bar{\Omega})$$
$$\quad (X^\top X\hat{\Omega} + \lambda I)^{-1}X^\top\big]$$
$$\quad - \mathrm{tr}\big[X(\hat{\Omega}X^\top X + \lambda I)^{-1}\Delta\hat{\Omega}X^\top X(\bar{\Omega}X^\top X + \lambda I)^{-1}$$
$$\quad (\bar{\Omega}X^\top X\bar{\Omega} + \lambda\bar{\Omega})(X^\top X\hat{\Omega} + \lambda I)^{-1}X^\top\big]$$
$$= \mathrm{tr}\big[X\bar{\Omega}(X^\top X\hat{\Omega} + \lambda I)^{-1}X^\top\big] - \mathrm{tr}\big[X(\hat{\Omega}X^\top X + \lambda I)^{-1}$$
$$\quad \Delta\hat{\Omega}X^\top X\bar{\Omega}(X^\top X\hat{\Omega} + \lambda I)^{-1}X^\top\big]$$
$$= \mathrm{tr}\big[X\bar{\Omega}(X^\top X\hat{\Omega} + \lambda I)^{-1}X^\top\big] - B/2.$$

Therefore we have

$$B + C - \mathrm{tr}\big[X\bar{\Omega}(X^\top X\hat{\Omega} + \lambda I)^{-1}X^\top\big]$$
$$= B/2 - \mathrm{tr}\big[X\bar{\Omega}(X^\top X\bar{\Omega} + \lambda I)^{-1}X^\top X\Delta\hat{\Omega}$$
$$\quad (X^\top X\hat{\Omega} + \lambda I)^{-1}X^\top\big]$$
$$= B/2 - \mathrm{tr}\big[X(\bar{\Omega}X^\top X + \lambda I)^{-1}\bar{\Omega}X^\top X\Delta\hat{\Omega}$$
$$\quad (X^\top X\hat{\Omega} + \lambda I)^{-1}X^\top\big]$$
$$= -\mathrm{tr}\big[X(\hat{\Omega}X^\top X + \lambda I)^{-1}\Delta\hat{\Omega}X^\top X$$
$$\quad (\bar{\Omega}X^\top X + \lambda I)^{-1}\bar{\Omega}X^\top X\Delta\hat{\Omega}(X^\top X\hat{\Omega} + \lambda I)^{-1}X^\top\big].$$

Putting all together, we have

$$A + B + C - \mathrm{tr}\big[X\bar{\Omega}(X^\top X\bar{\Omega} + \lambda I)^{-1}X^\top\big]$$
$$= \mathrm{tr}\big[X(\hat{\Omega}X^\top X + \lambda I)^{-1}\Delta\hat{\Omega}(I - X^\top X(\bar{\Omega}X^\top X + \lambda I)^{-1}$$
$$\quad \bar{\Omega})X^\top X\Delta\hat{\Omega}(X^\top X\hat{\Omega} + \lambda I)^{-1}X^\top\big]$$
$$= \lambda \mathrm{tr}\big[X(\hat{\Omega}X^\top X + \lambda I)^{-1}\Delta\hat{\Omega}(X^\top X\bar{\Omega} + \lambda I)^{-1}$$
$$\quad X^\top X\Delta\hat{\Omega}(X^\top X\hat{\Omega} + \lambda I)^{-1}X^\top\big].$$

This proves the upper bound. Similarly, the lower bound follows from the fact that $E\|X\hat{\beta} - X\bar{\beta}\|_2^2 \ge \sigma^2(A + B + C)$.